\documentclass[sigconf,10pt]{acmart}

\usepackage{algorithm}
\usepackage{algpseudocode}
\usepackage{amsmath, amsfonts}
\usepackage{graphicx}
\usepackage{booktabs}
\usepackage{lipsum} 
\usepackage{xcolor}    
\usepackage{listings}  
\newcommand{\systemname}{\texttt{SpecRouter}} 

\definecolor{backcolour}{rgb}{0.95,0.95,0.92}
\definecolor{codegreen}{rgb}{0,0.6,0}
\definecolor{codeblue}{rgb}{0,0,0.6} 
\definecolor{codegray}{rgb}{0.5,0.5,0.5}
\definecolor{codepurple}{rgb}{0.58,0,0.82}
\definecolor{codeorange}{rgb}{1.0,0.5,0} 

\lstdefinestyle{specrouterstyle}{
    backgroundcolor=\color{backcolour},
    commentstyle=\color{codegreen}\textit,
    keywordstyle=\color{codeblue}\bfseries, 
    numberstyle=\tiny\color{codegray},
    stringstyle=\color{codepurple},
    basicstyle=\footnotesize\ttfamily,
    breakatwhitespace=false,
    breaklines=true,
    captionpos=b,
    keepspaces=true,
    numbers=left,
    numbersep=5pt,
    showspaces=false,
    showstringspaces=false,
    showtabs=false,
    tabsize=2,
    emph={ChainRouter, Executor, StateManager, ModelPool, ModelChainScheduler, PerformanceProfiler, PrefillRequest, DraftRequest, VerifyRequest, RollbackRequest, ModelState, LogitsProcessorList},
    emphstyle=\color{codeorange}\bfseries, 
    morekeywords={Function, EndFunction, Require, Ensure, While, EndWhile, For, EndFor, If, EndIf, Else, Return, Break, True, False, new},
    identifierstyle=\color{black},
    morekeywords=[2]{Generate, GetOptimalChain, Execute, GetState, UpdateState, RecordMetrics, CheckTermination, ProcessVerificationStack, EstimateAcceptanceRate, PredictChainTime},
    keywordstyle=[2]{\color{codeorange}\bfseries},
    morecomment=[l]{//}, 
    escapeinside={\%*}{*)}, 
    literate=*{_}{{\textunderscore}}1 
}

\acmConference[XXXX '25]{ACM XXX}{May 2025}{XXX}
\acmYear{2025}
\acmISBN{978-1-4503-XXXX-X/25/10}

\setcopyright{acmcopyright}
\acmDOI{10.1145/1122445.1122456}

\title{SpecRouter: Adaptive Routing for Multi-Level Speculative Decoding in Large Language Models}

\author{Hang Wu} 
\authornote{These authors contributed equally to this work.}
\affiliation{%
  \institution{Xidian University}
  \city{Xian}
  \country{China}
}
\author{Jianian Zhu} 
\authornotemark[1]
\affiliation{%
  \institution{Huazhong University of Science and Technology}
  \city{Wuhan}
  \country{China}
}
\author{Yinghui Li}
\affiliation{%
  \institution{Tsinghua University}
  \city{Beijing}
  \country{China}
}
\author{Haojie Wang}
\affiliation{%
  \institution{Tsinghua University}
  \city{Beijing}
  \country{China}
}
\author{Biao Hou}
\affiliation{%
  \institution{Xidian University}
  \city{Xian}
  \country{China}
}
\author{Jidong Zhai}
\affiliation{%
  \institution{Tsinghua University}
  \city{Beijing}
  \country{China}
}

\begin{document}

\begin{abstract}
Large Language Models (LLMs) present a critical trade-off between inference quality and computational cost: larger models offer superior capabilities but incur significant latency, while smaller models are faster but less powerful. Existing serving strategies often employ fixed model scales or static two-stage speculative decoding, failing to dynamically adapt to the varying complexities of user requests or fluctuations in system performance. This paper introduces \systemname{}, a novel framework that reimagines LLM inference as an adaptive routing problem solved through multi-level speculative decoding. \systemname{} dynamically constructs and optimizes inference "paths" (chains of models) based on real-time feedback, addressing the limitations of static approaches. Our contributions are threefold: (1) An \textbf{adaptive model chain scheduling} mechanism that leverages performance profiling (execution times) and predictive similarity metrics (derived from token distribution divergence) to continuously select the optimal sequence of draft and verifier models, minimizing predicted latency per generated token. (2) A \textbf{multi-level collaborative verification} framework where intermediate models within the selected chain can validate speculative tokens, reducing the verification burden on the final, most powerful target model. (3) A \textbf{synchronized state management} system providing efficient, consistent KV cache handling across heterogeneous models in the chain, including precise, low-overhead rollbacks tailored for asynchronous batch processing inherent in multi-level speculation. Preliminary experiments demonstrate the validity of our method. 

\end{abstract}

\maketitle

\section{Introduction}

Large Language Models (LLMs) have become the driving force behind numerous artificial intelligence applications~\cite{touvron2023llama, grattafiori2024llama, chowdhery2023palm, liu2024deepseek}, from conversational systems to code generation and complex reasoning. However, the substantial growth in model scale leads to significant increases in computational demands and inference latency~\cite{zhou2024survey, fu2024serverlessllm, hu2024characterization, song2024powerinfer, sun2024llumnix}. Models with hundreds of billions of parameters can take seconds to generate a response, challenging real-time applications. The autoregressive nature of generation, where each token depends on previous ones~\cite{vaswani2017attention}, further exacerbates this by limiting parallelization, despite optimizations like KV caching~\cite{qin2025mooncake, zhu2025fastcache, lin2024infinite}.

LLM service providers face a dilemma: use large, high-quality but slow models, or smaller, faster but less capable ones~\cite{lin2024parrot, zhong2024distserve, wu2024loongserve, kaplan2020scaling}. This directly impacts user experience and service cost, especially under high concurrency. Existing strategies attempt to mitigate this trade-off but often fall short in adaptivity.

\textbf{Single-model services}~\cite{lin2024parrot, agrawal2024taming} deploy a fixed model size, leading to resource inefficiency for simple queries or suboptimal quality for complex ones. \textbf{Model cascades}~\cite{chen2024cascade, lebovitz2023efficient, yin2024theoretical} route requests through progressively larger models, but typically use static policies ill-suited for dynamic workloads. \textbf{LLM Routing}~\cite{zhang2025leveraging, feng2024graphrouter} directs queries to appropriate models, yet often relies on static rules or classifiers unable to adapt to real-time system conditions or query patterns.

\textbf{Speculative decoding (SD)}~\cite{leviathan2023fast, miao2024specinfer, li2024eagle} has emerged as a promising acceleration technique. It uses a smaller, faster "draft" model to propose tokens, which are then verified in parallel by the larger "target" model. While effective, current SD implementations typically employ a \textit{static two-stage} setup with a fixed draft-target pair. This approach suffers from several limitations. \textbf{Lack of Adaptivity:} Static pairs cannot adapt to the varying complexity of different requests or changes in system load and model performance over time. \textbf{Suboptimal Model Pairing:} Selecting the optimal draft model for a given target model is challenging. A naive choice (e.g., the smallest available model) can lead to high rejection rates due to significant distributional differences, negating the benefits of speculation~\cite{li2024eagle}. Finding the best pair often requires extensive empirical tuning, increasing deployment costs, especially with a large pool of models or diverse tasks. \textbf{Fixed Verification Strategy:} The verification process itself is typically fixed, without considering the possibility of intermediate verification steps or adapting the verification depth based on confidence or difficulty. These limitations highlight the need for a more dynamic and adaptive approach to speculative inference.


This paper introduces \systemname{}, a novel framework that addresses these shortcomings by reframing LLM inference as an adaptive routing problem solved via \textit{multi-level} speculative decoding. Instead of fixed pairs, \systemname{} dynamically constructs and optimizes inference "paths" – chains of models – based on real-time performance feedback and predictive metrics. It determines not only the optimal draft model(s) but also the optimal sequence of intermediate verifiers leading to the final target model.

Our key contributions are:
\begin{itemize} 

    \item \textbf{Adaptive Multi-Level Model Chain Scheduling:} We propose a dynamic scheduling mechanism that continuously selects the optimal chain of draft and verifier models from a pool of heterogeneous models. It uses real-time performance profiling and predictive similarity metrics (based on token distribution divergence) to minimize the predicted latency per effectively generated token, adapting the inference path to current conditions.
    \item \textbf{Collaborative Multi-Level Verification:} We introduce a framework where intermediate models within the selected chain collaboratively verify speculative tokens. This staged verification reduces the computational load on the most expensive target model by potentially rejecting incorrect tokens earlier using faster, intermediate verifiers.
    \item \textbf{Synchronized State Management for Heterogeneous Chains:} We design an efficient state management system specifically for multi-level, heterogeneous model chains. It ensures consistent KV cache handling and enables precise, low-overhead rollbacks across different models in the chain, crucial for managing the asynchronous progress inherent in batched multi-level speculation.
\end{itemize}

\systemname{} represents a significant step towards more efficient and adaptive LLM serving. By treating inference as a dynamically routed, collaborative process across multiple model levels, it enhances both individual request latency and overall system throughput, making high-quality LLM services more practical and responsive.

\section{Background}

\subsection{LLM Inference Fundamentals}

Large Language Models (LLMs) typically generate text autoregressively: each new token $x_t$ is sampled based on the preceding sequence $x_{<t}$. At each step $t$, the model computes hidden states, culminating in a probability distribution over the vocabulary $\mathcal{V}$:
\begin{equation}
p(x_t | x_{<t}) = \text{softmax}(h_t W)
\end{equation}
where $h_t$ is the final hidden state for position $t$, and $W$ is the output embedding matrix. This sequential dependency limits parallelism and becomes computationally expensive for large models.

\paragraph{KV Cache.} To mitigate redundant computations in the attention layers, the Key-Value (KV) cache stores the computed key ($K$) and value ($V$) tensors for previous tokens. For a model with $L$ layers, the cache at step $t$ contains $\{(K_i^{(<t)}, V_i^{(<t)}) | i=1...L\}$. When generating token $x_t$, only the $K, V$ tensors for the current token need to be computed and appended to the cache. While significantly reducing computation, the KV cache introduces substantial memory overhead, especially for long sequences and large batch sizes~\cite{kwon2023efficient}. Efficient management of this cache is critical.

\paragraph{Batch Processing.} Serving systems process multiple requests concurrently in batches to improve hardware utilization (throughput). However, managing batches in autoregressive generation is complex, as sequences within a batch can have different lengths and finish at different times. Techniques like continuous batching~\cite{kwon2023efficient} aim to handle this dynamically, but the fundamental memory bandwidth demands of loading model parameters and KV caches for large batches remain a bottleneck~\cite{sun2024llumnix, zhang2024fastlivemodelauto}.

\subsection{Speculative Decoding}

Speculative Decoding (SD) accelerates LLM inference by reducing the number of forward passes required through the large, slow target model ($M_p$). It uses a smaller, faster draft model ($M_q$) to generate a sequence of candidate tokens, which $M_p$ then verifies in parallel~\cite{leviathan2023fast, li2025eagle3scalinginferenceacceleration}.

\paragraph{Basic Speculative Decoding.} The process typically involves the following steps:
\begin{enumerate}
    \item The draft model $M_q$, given context $x_{<t}$, autoregressively generates a sequence of $\gamma$ candidate tokens: $\hat{x}_{t}, \hat{x}_{t+1}, ..., \hat{x}_{t+\gamma-1}$.
    \item The target model $M_p$ performs a single forward pass, taking $x_{<t}$ and the drafted sequence as input, to compute the true probability distributions $p(x_{t+i} | x_{<t+i})$ for $i \in [0, \gamma-1]$.
    \item The drafted tokens are verified sequentially. Token $\hat{x}_{t+i}$ is accepted if it matches the token sampled from $p(x_{t+i} | x_{<t+i})$ according to a chosen sampling rule (e.g., greedy, or probabilistic acceptance~\cite{leviathan2023fast}). Verification stops at the first rejection.
    \item If $k$ tokens are accepted ($0 \le k \le \gamma$), the target model generates one additional token $x_{t+k}$ based on $p(x_{t+k} | x_{<t+k})$. The final accepted sequence for this step is $x_t, ..., x_{t+k}$.
\end{enumerate}
This allows generating up to $\gamma+1$ tokens with only one forward pass of $M_p$ and $\gamma$ passes of $M_q$.

\paragraph{Efficiency Analysis.} The effectiveness of SD depends on the trade-off between the draft model's speed and its prediction accuracy (acceptance rate). Let $p(x)$ and $q(x)$ be the probability distributions of the target and draft models, respectively.
The theoretical acceptance probability $\alpha$ (probability that a drafted token is accepted) can be related to the similarity between the distributions, often approximated as~\cite{leviathan2023fast, li2024eagle}:
\begin{equation}
\alpha \approx \mathbb{E}\left[\sum_{x \in \mathcal{V}} \min(p(x), q(x))\right] = 1 - \mathbb{E}[\text{DTV}(p, q)]
\end{equation}
where DTV is the Total Variation Distance (defined later in Eq.~\ref{eq:dtv}).
The expected number of tokens accepted from a draft sequence of length $\gamma$ is approximately~\cite{leviathan2023fast}:
\begin{equation}
E[\text{accepted\_tokens}] \approx \frac{1 - \alpha^{\gamma+1}}{1 - \alpha}
\end{equation}
Considering the single extra token generated by $M_p$, the expected number of tokens generated per $M_p$ forward pass is $E[\text{total\_tokens}] \approx 1 + E[\text{accepted\_tokens}]$.
Let $T_p$ and $T_q$ be the per-token latency of the target and draft models, respectively, and let $c = T_q / T_p$ be their relative cost. The latency of one SD step is approximately $\gamma T_q + T_p = (\gamma c + 1) T_p$. The theoretical speedup (Improvement Factor) over standard autoregressive decoding (latency $T_p$ per token) is:
\begin{equation} \label{eq:sd_speedup}
\text{Speedup} = \frac{E[\text{total\_tokens}] \times T_p}{\text{Latency per SD step}} \approx \frac{1 + \sum_{i=1}^{\gamma} \alpha^i}{(\gamma c + 1)} = \frac{1 - \alpha^{\gamma+1}}{(1 - \alpha)(\gamma c + 1)}
\end{equation}
This shows that speedup depends critically on achieving a high acceptance rate $\alpha$ with a fast draft model (low $c$) and an appropriate draft length $\gamma$.

\paragraph{Limitations of Existing SD} While promising, achieving consistent speedup with basic SD is challenging. \textbf{Distribution Mismatch:} If $M_q$ is much smaller or trained differently than $M_p$, their output distributions $p(x)$ and $q(x)$ can diverge significantly, leading to a low acceptance rate $\alpha$ and poor speedup. Finding a good pair often requires costly fine-tuning (distillation) or extensive empirical search~\cite{zhou2023distillspec,li2025eagle3scalinginferenceacceleration}. \textbf{Static Configuration:} Most systems use a fixed $(M_q, M_p, \gamma)$ configuration. This is suboptimal as the ideal $\gamma$ depends on $\alpha$ and $c$, and both $\alpha$ and the relative cost $c$ can fluctuate based on the input context (e.g., coding vs. prose generation) and system load. \textbf{Two-Stage Limitation:} The fixed two-stage structure doesn't allow for intermediate verification steps, potentially missing opportunities for earlier rejection by a moderately sized model, which could be faster than waiting for the final target model. Recent works like tree-based SD (e.g., SpecInfer~\cite{miao2024specinfer}, Medusa~\cite{cai2024medusa}) improve drafting by exploring multiple candidates in parallel, increasing the chance of acceptance. However, they still typically rely on static configurations and don't address the adaptive multi-level challenge. \systemname{} aims to overcome these limitations through dynamic, multi-level chain selection and verification.



\section{System Overview}

\subsection{Design Goals and Principles}

\systemname{} tackles the LLM inference efficiency challenge by drawing inspiration from adaptive network routing. It manages speculative decoding across a pool of heterogeneous models by dynamically optimizing the inference path. Key principles guide its design: \textbf{Adaptive Path Optimization:} Continuously select the most efficient sequence (chain) of models for drafting and verification based on real-time performance and model characteristics, mimicking dynamic routing protocols. \textbf{Latency Minimization under Quality Constraints:} Maximize acceleration while rigorously preserving the output quality equivalent to that of the designated final target model. \textbf{Efficient State Synchronization:} Provide robust, low-overhead mechanisms for managing KV cache consistency across heterogeneous models in the chain, especially during rollbacks required by speculation. \textbf{Resource Efficiency and Scalability:} Effectively manage diverse model resources (CPU/GPU memory, compute) and scale to handle concurrent requests efficiently. These principles translate into concrete design goals: implementing dynamic model chain scheduling, providing a robust multi-level verification framework, ensuring efficient state management with atomic rollbacks, and maintaining a modular architecture.

\subsection{System Architecture}
\label{sec:architecture_overview}

\begin{figure} 
\centering
\includegraphics[width=0.9\columnwidth]{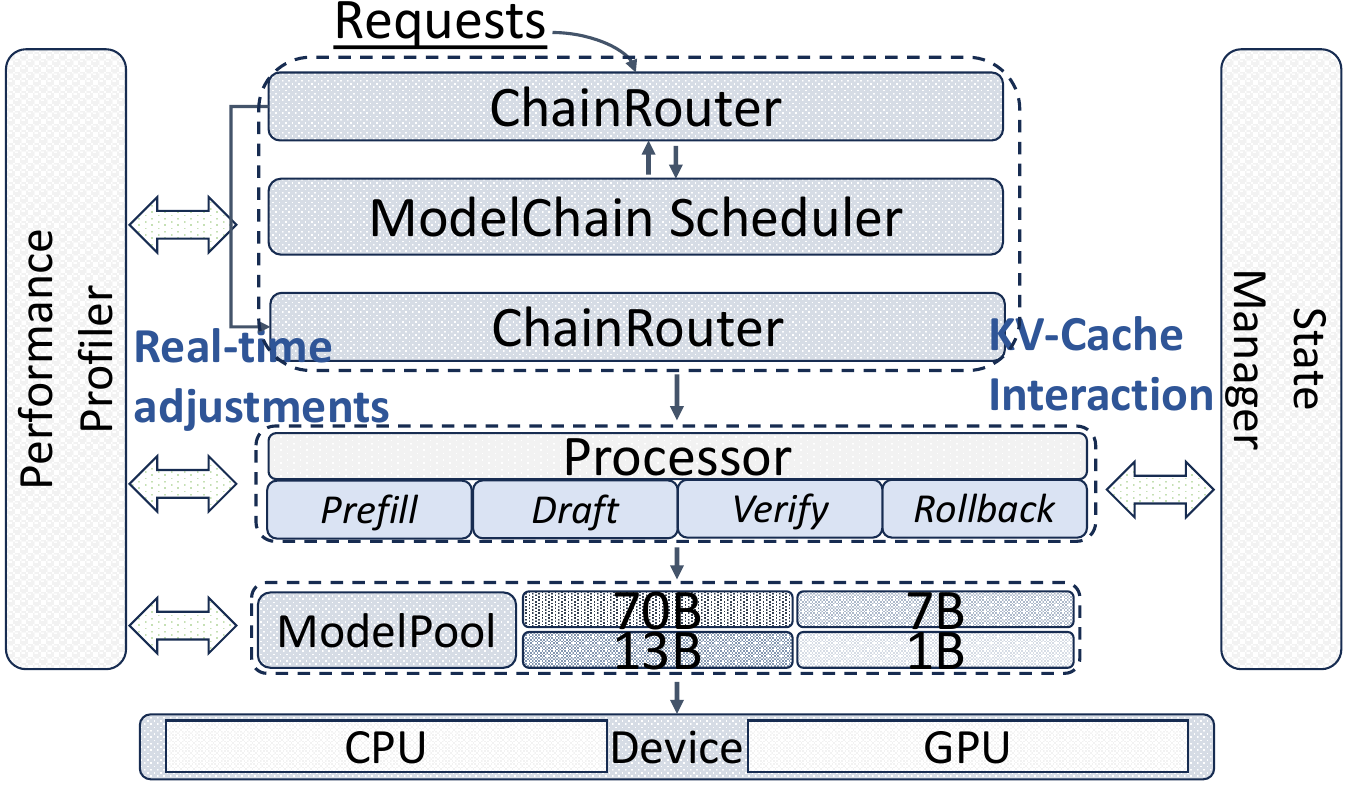} 
\caption{Architecture of \systemname{}}
\label{fig:system_architecture}
\end{figure}

\systemname{} employs a modular architecture, illustrated in Figure~\ref{fig:system_architecture}, comprising several key components. The central control plane is the \textbf{\texttt{ChainRouter}}; it receives generation requests, orchestrates the multi-level speculative decoding workflow (prefill, draft, verify, rollback), interacts with the \texttt{ModelChainScheduler} to get optimal model chains, dispatches tasks to the \texttt{Executor}, and manages the overall generation state and termination logic. Resource management is handled by the \textbf{\texttt{ModelPool}} and \textbf{\texttt{DeviceManager}} layer. This layer manages the lifecycle (registration, loading, caching, allocation, garbage collection) of heterogeneous models across available devices (GPUs, CPU) and provides models to the execution layer on demand. The \textbf{\texttt{Executor}} serves as the primary data plane dispatcher, receiving operational requests from the \texttt{ChainRouter} and routing them to specialized, stateless \textbf{\texttt{Processors}} (\texttt{Prefill-}, \texttt{Draft-}, \texttt{Verify-}, \texttt{RollbackProcessor}). These processors are workers executing specific stages of the speculative process, interacting with the \texttt{ModelPool} for computation and the \texttt{StateManager} for state. The \textbf{\texttt{StateManager}} manages persistent inference state, primarily the KV caches (\texttt{ModelState}) associated with ongoing requests, offering mechanisms for state creation, retrieval, consistent updates, and atomic rollbacks crucial for multi-level speculation. The adaptive intelligence core is the \textbf{\texttt{ModelChainScheduler}}, analyzing real-time performance data to dynamically select the optimal chain of models for subsequent draft/verify steps to minimize predicted latency per token. Monitoring is performed by the \textbf{\texttt{PerformanceProfiler}}, the monitoring plane, which gathers low-overhead timing and counter metrics across all operations, providing the feedback loop necessary for the \texttt{ModelChainScheduler}'s adaptive decisions.

In operation, the \texttt{ChainRouter}, guided by the \texttt{ModelChainScheduler}, directs the speculative flow. It dispatches tasks via the \texttt{Executor} to the appropriate \texttt{Processors}. These processors use models from the \texttt{ModelPool} and manage context via the \texttt{StateManager}. The \texttt{PerformanceProfiler} observes this activity, feeding metrics back to the \texttt{ModelChainScheduler}, which updates its routing strategy (the optimal model chain) for subsequent steps, closing the adaptive loop.

\subsection{Key Design Insight: Inference as Adaptive Multi-Hop Routing}

\systemname{} differentiates itself by conceptualizing LLM inference as a data flow requiring efficient, adaptive routing across heterogeneous model nodes, inspired by network routing principles. \textbf{Dynamic Multi-Hop Inference Routing:} Unlike static two-stage SD, \systemname{} implements a generalized multi-level model chain framework. Analogous to dynamic routing protocols, it discovers and constructs optimal inference paths (model chains $[M_k, ..., M_t]$) of varying depths based on real-time conditions (model performance, similarity). The final target model $M_t$ acts as the ultimate verifier, ensuring quality. \textbf{State-Centric Session Management:} The KV cache is treated as the core state for inference session continuity, similar to connection state in routers. \systemname{} implements efficient state synchronization, reuse, and rollback mechanisms across the heterogeneous models in the chain, ensuring consistency as the inference flow traverses different nodes. \textbf{Adaptive Routing and Flow Control:} Runtime feedback (latency, acceptance rates) from draft and verification stages informs continuous adaptation, mirroring network feedback mechanisms. \systemname{} dynamically adjusts its routing table (model chain selection) and speculative parameters (e.g., effective window size) to optimize throughput and latency. This network-inspired adaptive routing approach enables \systemname{} to achieve significant performance gains by dynamically tailoring the inference process to the specific request and current system state.

\section{System Design and Details}

\systemname{} orchestrates multi-level speculative decoding as an intelligent routing system. It aims to minimize latency while preserving the target model's output quality by dynamically selecting and executing inference paths (model chains). The core control flow (Listing~\ref{lst:specrouter_core_flow}) involves iteratively scheduling an optimal chain, drafting tokens, verifying them potentially across multiple levels, and synchronizing state. This section details the components enabling this adaptive process.

\begin{lstlisting}[style=specrouterstyle, caption={Core Control Flow of the \systemname{} Generation Loop}, label={lst:specrouter_core_flow}, language=Python]
# High-level conceptual representation
def core_generate(prompt, max_tokens, target_model_id):
    # 1. Initialize state, prefill prompt using target model & potentially others
    current_engine_state = Initialize_And_Prefill(prompt, target_model_id)
    generated_token_count = 0
    is_generation_complete = False
    # 2. Iteratively generate tokens
    while not is_generation_complete:
        # 2a. Get dynamically optimized model chain from scheduler
        # Input: current state, target_model_id, performance metrics
        # Output: optimal_model_chain = [M_draft, ..., M_intermediate, ..., M_target]
        optimal_model_chain = ModelChainScheduler.Get_Optimal_Chain(current_engine_state, target_model_id)
        # 2b. Execute one multi-level speculative step using the chosen chain
        step_outcome = Executor.Execute_Speculative_Step(optimal_model_chain,
                       current_engine_state)
        # 2c. Update engine state and token count
        current_engine_state = step_outcome.get_updated_engine_state()
        num_accepted = len(step_outcome.get_accepted_tokens()) # Pseudocode
        generated_token_count += num_accepted
        # 2d. Check termination conditions (EOS, max_tokens, errors)
        is_generation_complete = Check_Termination_Condition(current_engine_state, generated_token_count, max_tokens)
    # 3. Extract final result
    final_output = Extract_Final_Result(current_engine_state)
    return final_output
\end{lstlisting}

\subsection{The \texttt{ChainRouter}: Central Coordination}
\label{sec:chainrouter}

The \texttt{ChainRouter} acts as the control plane, orchestrating the inference lifecycle based on scheduler guidance and system state.

\paragraph{Initialization and Configuration.} Upon receiving a request, it tokenizes the prompt using the target model's ($M_t$) tokenizer (obtained from \texttt{ModelPool}), sets up sampling parameters (\texttt{LogitsProcessorList}), and defines generation constraints (\texttt{max\_new\_tokens}, \texttt{eos\_token\_id}).

\paragraph{Prefill Phase.} It initiates the prefill process. It may prefill using only the target model $M_t$ or potentially prefill across an initial chain recommended by the scheduler. It constructs \texttt{PrefillRequest} messages containing the tokenized prompt and dispatches them via the \texttt{Executor}. The results include initial \texttt{ModelState} identifiers (managed by \texttt{StateManager}) and potentially initial logits used by the scheduler for baseline similarity calculations. Prefill failure, especially on $M_t$, typically terminates the request.

\paragraph{Iterative Speculation Loop.} In each generation cycle:
\begin{enumerate}
    \item \textbf{Get Optimal Chain:} Queries the \texttt{ModelChainScheduler} (\S~\ref{sec:scheduler}) for the best model chain $\texttt{Chain}_{opt} = [M_{k_1}, ..., M_t]$ based on current metrics.
    \item \textbf{Dispatch Operations:} Sequentially dispatches tasks via the \texttt{Executor}:
        \begin{itemize}
            \item A \texttt{DraftRequest} to the first model $M_{k_1}$.
            \item \texttt{VerifyRequest} messages to subsequent models $M_{k_2}, ..., M_t$ in the chain.
            \item Potentially \texttt{RollbackRequest} messages based on verification outcomes at different levels.
        \end{itemize}
        It tracks the \texttt{state\_id} for each model throughout this process.
    \item \textbf{Update State:} Updates the common input context (\texttt{engine\_inputs}) for the next draft based on the tokens accepted in the current step.
    \item \textbf{Check Termination:} Evaluates stopping conditions (EOS reached, \texttt{max\_new\_tokens} exceeded) for all sequences in the batch based on the final synchronized state.
\end{enumerate}

\paragraph{Error Handling.} If the \texttt{Executor} reports a failure during a step (e.g., OOM), the \texttt{ChainRouter} attempts recovery, potentially by requesting a more robust (e.g., shorter or simpler) chain from the scheduler and retrying. Persistent failures, especially on $M_t$, lead to graceful request termination.

\subsection{The \texttt{ModelChainScheduler}: Dynamic Model Chain Scheduling}
\label{sec:scheduler}

The \texttt{ModelChainScheduler} implements \systemname{}'s core adaptivity, dynamically selecting the optimal model chain $[M_1, M_2, ..., M_N]$ (where $M_1$ is the initial draft, $M_N$ is the final target $M_t$) to minimize the predicted effective latency per generated token.

\paragraph{Input Metrics.} Decisions are based on real-time metrics updated via EMA (Exponential Moving Average) using data from the \texttt{PerformanceProfiler}. \textbf{$T_{i}$}: Smoothed average per-token execution time for model $M_i$, calculated as $T_{i}^{\text{new}} = \alpha_{\text{time}} T_{i}^{\text{measured}} + (1 - \alpha_{\text{time}}) T_{i}^{\text{old}}$. \textbf{$\text{SimScore}(M_i, M_j)$}: Predictive similarity between models $M_i$ (draft) and $M_j$ (verifier), calculated using the average Total Variation Distance (DTV) between their output distributions $p_{M_i}$ and $p_{M_j}$ over the vocabulary $\mathcal{V}$:
\begin{equation} \label{eq:dtv}
\text{DTV}(p_{M_i}, p_{M_j}) = \frac{1}{2} \sum_{v \in \mathcal{V}} |p_{M_i}(v) - p_{M_j}(v)|
\end{equation}
DTV is chosen for its symmetry and computational efficiency. The similarity score, also updated via EMA, is:
\begin{equation} \label{eq:simscore}
\text{SimScore}(M_i, M_j) = 1 - \mathbb{E}[\text{DTV}(p_{M_i}, p_{M_j})]
\end{equation}
where $\mathbb{E}[\cdot]$ is the smoothed average DTV. \textbf{$\alpha_{ij}$}: The effective acceptance probability when $M_j$ verifies tokens drafted by $M_i$, estimated from the similarity score: $\alpha_{ij} \approx f(\text{SimScore}(M_i, M_j))$, where $f$ is a mapping function (e.g., calibrated sigmoid). \textbf{$W$}: The speculative draft window size (number of tokens drafted by $M_1$), typically configured globally.

\paragraph{Chain Efficiency Prediction.} The core task is to predict the effective time per target token ($T_{\text{eff}}$) for a candidate chain $\mathcal{C} = [M_1, ..., M_N=M_t]$. This involves modeling the expected cost considering draft time, verification times at each level, and the probabilities of reaching each level. Let $T_i$ be the per-token time for $M_i$, and $\alpha_{i, i+1}$ be the acceptance probability when $M_{i+1}$ verifies $M_i$'s draft of size $W$. The expected number of tokens accepted by $M_{i+1}$ from $M_i$'s draft is $E[\text{acc}_{i,i+1}] \approx \sum_{k=1}^{W} (\alpha_{i,i+1})^k$. The probability of needing verification by $M_{i+1}$ depends on acceptance at prior levels. A simplified model for the effective time per target token, $T_{\text{eff}}(\mathcal{C})$, can be expressed as the total expected time divided by the expected number of target tokens generated per cycle:
\begin{equation} \label{eq:cascade_time_revised}
T_{\text{eff}}(\mathcal{C}) \approx \frac{\text{Expected Latency per Cycle}}{\text{Expected Target Tokens per Cycle}}
\end{equation}
The numerator involves summing the costs: $W \times T_1$ (drafting) plus the expected verification costs at levels $M_2, ..., M_N$. The verification cost at level $M_j$ ($j>1$) is roughly $W \times T_j$ multiplied by the probability that verification reaches level $j$. The denominator is the expected number of tokens finally accepted by $M_N$. This prediction balances the speed of early models ($T_i$) against the likelihood ($\alpha_{ij}$) of their outputs being accepted by later models. The function $\texttt{Predict\_Effective\_Time}$ in Algorithm~\ref{alg:chain_opt_simplified} implements this prediction logic.

\paragraph{Chain Selection Algorithm (Algorithm~\ref{alg:chain_opt_simplified}).}
\begin{enumerate}
    \item \textbf{Generate Candidates:} Identify potential model chains $\mathcal{C}_{\text{candidate}}$ ending with the target model $M_t$.
    \item \textbf{Predict Efficiency:} For each candidate chain $\mathcal{C}$, estimate $T_{\text{eff}}(\mathcal{C})$ using $\texttt{Predict\_Effective\_Time}(\mathcal{C}, T, \alpha, W)$ based on current metrics.
    \item \textbf{Select Optimum:} Choose the chain $\text{Chain}_{\text{opt}}$ with the minimum predicted $T_{\text{eff}}$.
\end{enumerate}
This process repeats, allowing dynamic adaptation of the inference path.

\begin{figure}[ht] 
    \centering
    \includegraphics[width=0.9\columnwidth]{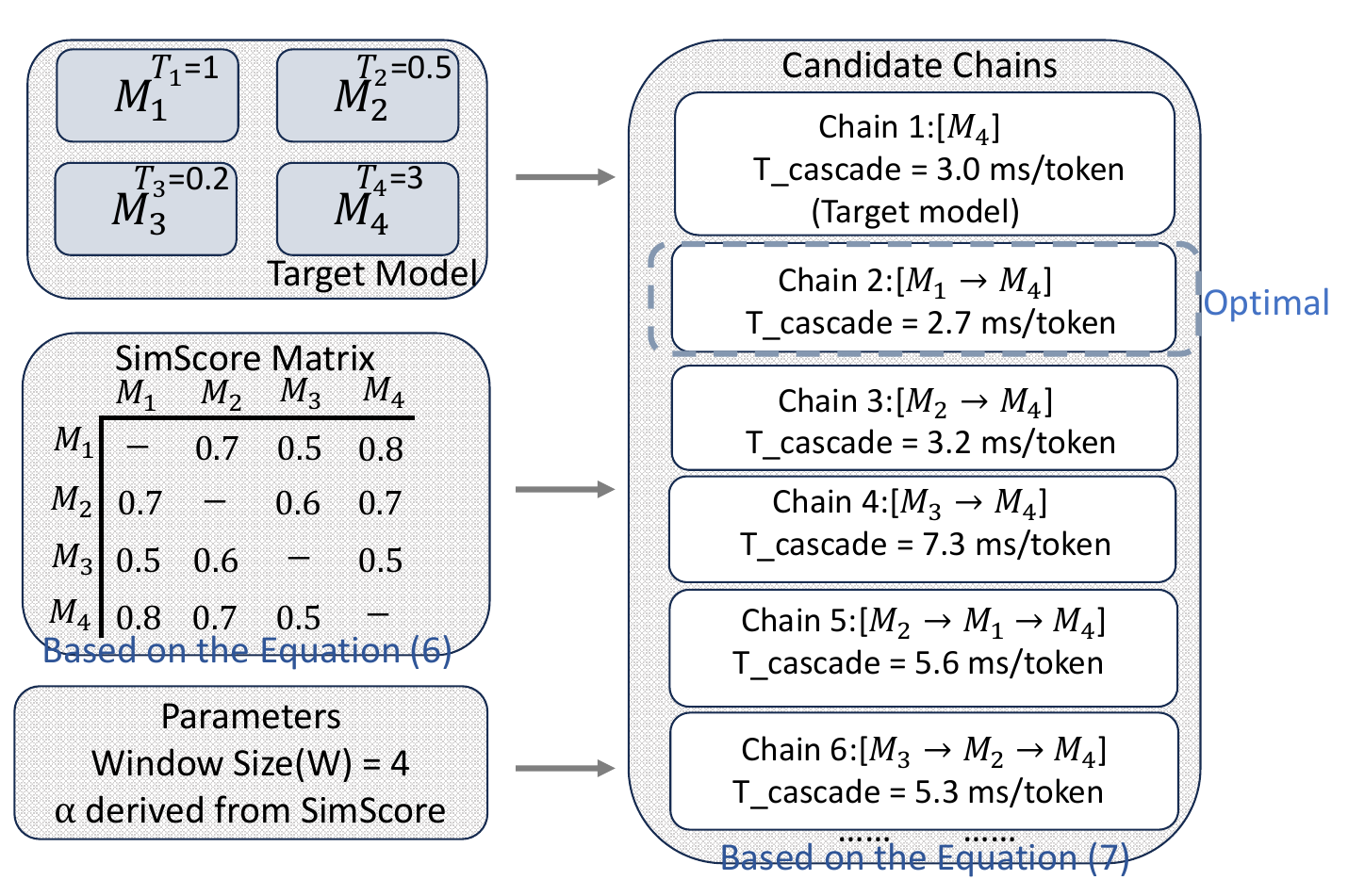} 
    \caption{Dynamic model chain selection example. Based on profiled times ($T_i$) and acceptance probabilities ($\alpha_i$, derived from SimScore), the scheduler predicts the effective time per target token ($T_{\text{eff}}$) for candidate chains (using logic based on Eq.~\ref{eq:cascade_time_revised}, conceptualized in Alg.~\ref{alg:chain_opt_simplified}'s prediction function) and selects the chain with the minimum predicted time.}
    \label{fig:scheduler_revised}
\end{figure}

\begin{algorithm}[ht]
\caption{Optimal Model Chain Selection}
\label{alg:chain_opt_simplified}
\begin{algorithmic}[1]
  \Require Set of available models $\mathcal{M}$; target model $M_t$; window size $W$
  \Require Current performance metrics: $T = \{T_i\}$ (times), $S = \{S_{ij}\}$ (similarities)
  \Ensure Optimal model chain $\text{Chain}_{\text{opt}}$

  \State \Comment{Step 1: Generate candidate chains ending with $M_t$}
  \State sorted\_models $\leftarrow$ \textsc{SortModelsByCapability}($\mathcal{M}$) 
  \State $\mathcal{C}_{\text{candidate}} \leftarrow$ \textsc{GenerateCandidateChains}(sorted\_models, $M_t$)
  \State \Comment{e.g., {[$M_k$, \ldots, $M_t$]}, {[$M_j$, \ldots, $M_t$]}, \ldots, {[$M_t$]}}

  \State \Comment{Step 2: Find chain with minimum predicted effective time}
  \State $T_{\min\_eff} \leftarrow \infty$
  \State $\text{Chain}_{\text{opt}} \leftarrow [M_t]$ \Comment{Default to target-only}

  \ForAll{$\text{chain} \in \mathcal{C}_{\text{candidate}}$}
    \State \Comment{Estimate acceptance probs $\alpha_{ij}$ from similarities $S_{ij}$}
    \State $\alpha \leftarrow \textsc{EstimateAcceptanceProb}(\text{chain}, S)$
    \State \Comment{Predict effective time per target token for this chain}
    \State $T_{\text{predicted\_eff}} \leftarrow \textsc{Predict\_Effective\_Time}(\text{chain}, T, \alpha, W)$ \Comment{Implements Eq.~\ref{eq:cascade_time_revised} logic}
    \If{$T_{\text{predicted\_eff}} < T_{\min\_eff}$}
      \State $T_{\min\_eff} \leftarrow T_{\text{predicted\_eff}}$
      \State $\text{Chain}_{\text{opt}} \leftarrow \text{chain}$
    \EndIf
  \EndFor

  \State \Return $\text{Chain}_{\text{opt}}$
\end{algorithmic}
\end{algorithm}

\subsection{The \texttt{Processors} and Execution Stages}
\label{sec:execution_processors}

The \texttt{Executor} routes requests to specialized processors for each stage:

\paragraph{\texttt{PrefillProcessor}.} Handles initial prompt processing, populates initial \texttt{ModelState} (KV cache, tokens, mask) via \texttt{StateManager}, and returns initial logits/state IDs.

\paragraph{\texttt{DraftProcessor}.} Executes speculative generation using the draft model and its state (KV cache) to generate $W$ tokens. Updates draft state and returns proposals.

\paragraph{\texttt{VerifyProcessor}.} Performs verification at a specific level. Aligns state, runs verifier forward pass, compares logits with proposals, determines accepted prefix, updates verifier's \texttt{ModelState}, and returns results. Repeats for each verification level in the chain.

\paragraph{\texttt{RollbackProcessor}.} Ensures state consistency post-verification. Calculates rollback length for each model based on consensus, invokes \texttt{StateManager} to atomically truncate KV caches and associated state (\texttt{cache\_tokens}, \texttt{cache\_mask}).

\subsection{The \texttt{StateManager} for Multi-Level State Consistency}
\label{sec:statemanagement}

Managing KV cache consistency across a dynamic chain of heterogeneous models with asynchronous batch progress is critical.

\paragraph{Challenge: Asynchronous Progress.} Different sequences in a batch might accept different numbers of speculative tokens, creating divergent states that complicate KV cache management.

\begin{figure}[ht] 
    \centering
    \includegraphics[width=0.9\columnwidth]{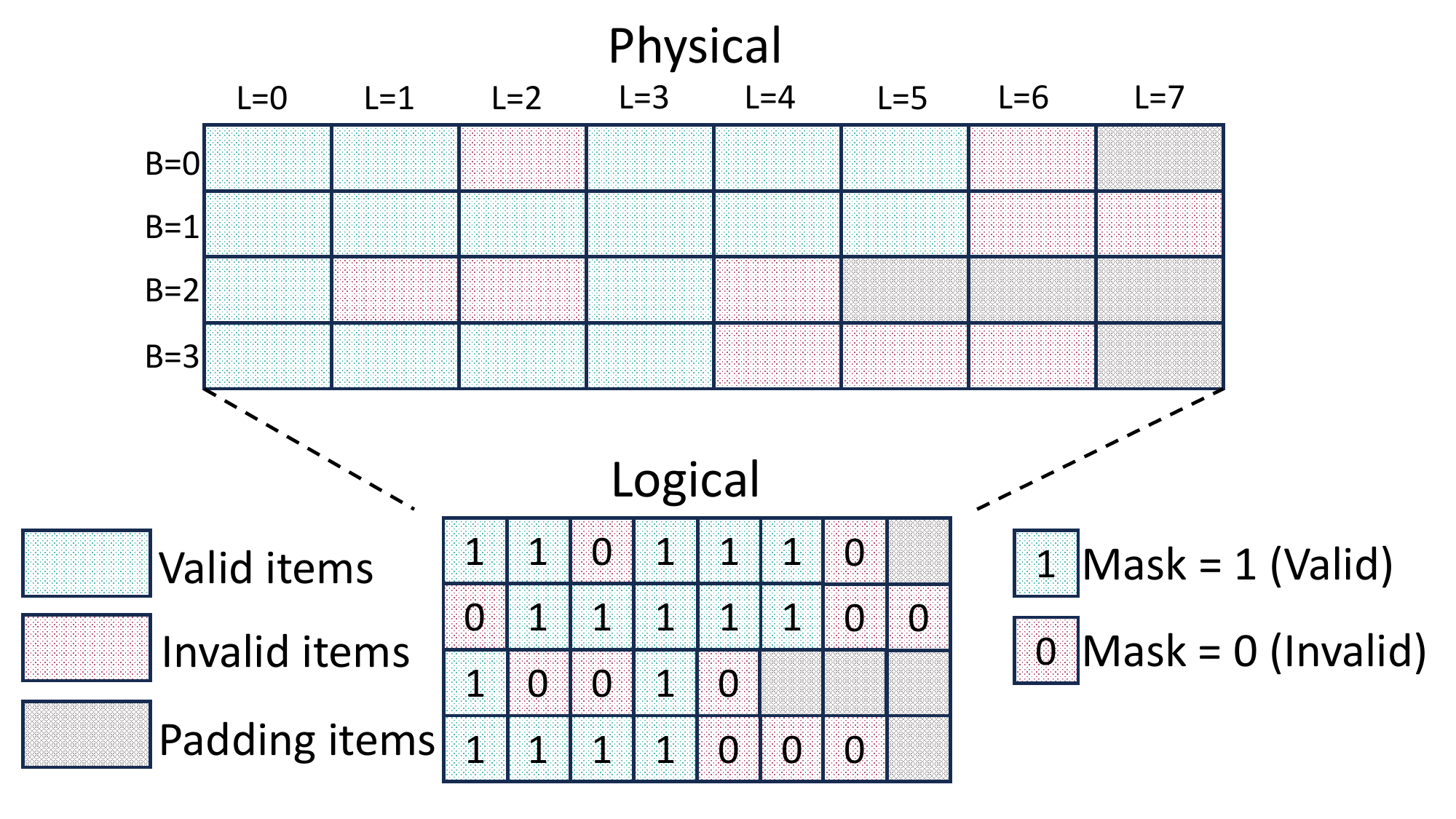}
    \caption{Illustration of the logical \texttt{cache\_mask} decoupling validity from physical KV cache storage. Invalid items (Mask=0) are ignored during attention, enabling correct handling of asynchronous progress after rollbacks without immediate data movement.}
    \label{fig:kv_mask}
\end{figure}

\paragraph{Solution: Logical Validity Mask.} The \texttt{StateManager} uses a \texttt{ModelState} abstraction containing the physical KV cache, token IDs (\texttt{cache\_tokens}), and a crucial logical validity mask, \texttt{cache\_mask} $\in \{0,1\}^{B \times L}$ (Figure~\ref{fig:kv_mask}). This mask tracks logically valid entries, decoupling validity from physical storage. An entry marked 0 is ignored by attention, even if data exists.

\paragraph{Rollback Mechanism.} When sequence $b$ needs to roll back $r_b$ tokens:
\begin{enumerate}
    \item \textbf{Logical Rollback:} The \texttt{RollbackProcessor} immediately updates \texttt{cache\_mask} for sequence $b$, setting the last $r_b$ entries to 0. The attention mechanism in the next step uses this mask to ignore invalid entries (Eq.~\ref{eq:attn_mask_revised}), ensuring correctness without immediate data movement.
    \item \textbf{Physical Truncation (Optimization):} If the minimum rollback length across the batch ($r_{\min} = \min_{b} r_b$) is greater than 0, the \texttt{StateManager}'s \texttt{fix\_kv\_cache} method physically truncates the trailing $r_{\min}$ entries from all relevant tensors (Eq.~\ref{eq:kv_rollback_revised}), reclaiming common memory.
\end{enumerate}
\begin{equation} \label{eq:attn_mask_revised}
\text{AttentionMask}_b = \text{GenerateAttentionMask}(\text{\texttt{cache\_mask}}[b, :L'_b])
\end{equation}
\begin{equation} \label{eq:kv_rollback_revised}
\text{Tensor}_{\text{new}} = \text{Tensor}_{\text{old}}[..., :(L - r_{\min}), ...] \quad (\text{if } r_{\min} > 0)
\end{equation}
where $L'_b$ is the new logical length after rollback, and $L$ is the previous physical length.

This hybrid approach ensures consistency and performance. The \texttt{StateManager} also handles state lifecycle, concurrency, and garbage collection.

\subsection{Resource Management (\texttt{ModelPool} and \texttt{DeviceManager})}
\label{sec:modelpool}

The \texttt{ModelPool} and \texttt{DeviceManager} manage heterogeneous model resources. They handle model lifecycle (registration, lazy loading, caching, garbage collection) and allocate models to appropriate devices (GPU/CPU) based on availability and requirements, abstracting these details from the execution logic.

\subsection{Performance Profiling (\texttt{PerformanceProfiler}) and Feedback Loop}
\label{sec:profiling}

The \texttt{PerformanceProfiler} gathers low-overhead metrics like execution times ($T_i^{\text{measured}}$) and token acceptance counts. This data feeds the \texttt{ModelChainScheduler}, enabling adaptive chain selection by providing updated estimates of model latencies ($T_i$) and inter-model acceptance probabilities ($\alpha_{ij}$ derived from SimScore).

\subsection{Fault Tolerance and Resilience}
\label{sec:fault_tolerance} 

\systemname{} includes resilience features. The \texttt{Executor} catches processor errors, allowing the \texttt{ChainRouter} to attempt recovery (e.g., requesting a different chain). The \texttt{ModelPool} offers CPU fallback for resource constraints. These enhance robustness.

\section{Implementation}
\label{sec:implementation}
\paragraph{\underline{Core Implementation and Dependencies}}
\systemname{} is implemented primarily in Python, comprising approximately 16K lines of code\footnote{Estimated based on core modules.}, leveraging the PyTorch library~\cite{paszke2019pytorch} for tensor operations and GPU acceleration, and the Hugging Face Transformers library~\cite{wolf2020transformers} for foundational LLM loading and manipulation. The system architecture emphasizes modularity, allowing for independent development and testing of its core components like scheduling, execution, and state management. 
\paragraph{\underline{Multi-GPU Model Deployment}}
\systemname{} is designed to operate efficiently on a single compute node equipped with multiple GPUs, such as the target platform featuring 10 NVIDIA A100 GPUs~\cite{choquette2020nvidia}. The deployment and management of the heterogeneous model pool across these GPUs are handled by the \texttt{ModelPool} in conjunction with a \texttt{DeviceManager}. Rather than implementing fine-grained model parallelism techniques like Tensor Parallelism (TP) to split individual large models or classical Pipeline Parallelism (PP) across micro-batches, our current approach focuses on the strategic placement of *entire, distinct models* onto specific GPU devices. Each assistant model and the target model instance are treated as independent units. The \texttt{DeviceManager} assigns each required model to an available GPU, considering factors like memory capacity and current load. This spatial distribution allows multiple models – potentially involved in different stages of the dynamically selected inference chain (e.g., a draft model on GPU 0, an intermediate model on GPU 1, and the final target model on GPU 2) – to reside concurrently in GPU memory. This facilitates rapid switching between models during the execution of a speculative chain managed by the \texttt{Executor} and minimizes interference between models. While underlying libraries might employ TP for loading extremely large individual models if necessary, the primary multi-GPU strategy within \systemname{} itself revolves around this device-level placement and scheduling of separate model instances across the available GPUs on the node.
\paragraph{\underline{Models}}
\label{sec:exp_models}
A key objective of our evaluation is to demonstrate \systemname{}'s ability to automatically and efficiently leverage a~\texttt{ModelPool} of heterogeneous language models. Our experiments utilize models exclusively from the Llama family~\cite{touvron2023llama, grattafiori2024llama}, including variants like \texttt{Llama-2-7b}, \texttt{Llama-2-13b}, \texttt{TinyLlama-1.1B}, and \texttt{Llama-68m}. We focus on this family due to its architectural representativeness and, critically, the availability of a consistent tokenizer across different model sizes. 

During experiments, a specific model from this pool (e.g., \texttt{Llama-2-70b}) is designated as the final target model ($M_t$), guaranteeing the desired output quality. All other compatible models residing in the \texttt{ModelPool} are then automatically available to the \texttt{ModelChainScheduler}. The scheduler dynamically selects from these available models to construct the most efficient multi-level speculative inference chain (including drafters and intermediate verifiers) in real-time, based on runtime performance metrics and model similarity scores, without requiring manual pre-configuration of specific draft-verifier pairs. This automatic chain formation and optimization capability is central to \systemname{}'s design. All models are loaded and computations are performed using the \texttt{bfloat16} data type for optimal performance on our target hardware.
\paragraph{\underline{Baselines}}
\label{sec:exp_baselines}
To rigorously evaluate the performance benefits of \systemname{}, we compare it against several key baselines:
\begin{itemize}
    \item \textbf{Target Model Only (TMO):} This baseline utilizes only the designated target model performing standard autoregressive decoding without any speculation. It serves as the fundamental reference for output quality (correctness) and represents the performance lower bound (in terms of throughput and latency) that speculative methods aim to surpass.
\item \textbf{Static Speculative Decoding (SSD):} This represents conventional speculative decoding approache~\cite{leviathan2023fast, miao2024specinfer}using a fixed configuration. We specifically evaluate:
        \begin{itemize}
            \item \textit{SSD-Smallest:} A common naive strategy where the smallest available assistant model in the pool (e.g., \texttt{Llama-68m} when the target is \texttt{Llama-2-7b}) is statically chosen as the draft model ($M_q$). A fixed draft length ($\gamma$) is used throughout the generation, typically requiring tuning for optimal performance.
            \item \textit{SSD-Tuned:} Represents an optimized static baseline where the best fixed pair $(M_q, M_t)$ and optimal fixed draft length $\gamma$ are pre-determined through extensive offline profiling or heuristics (e.g., based on pre-computed model similarity). While challenging to find the absolute optimum, this conceptual baseline highlights the performance achievable by a well-tuned static system, against which \systemname{}'s adaptivity can be compared.
        \end{itemize}
\end{itemize}
\paragraph{\underline{Workloads}}
Similar to prior work~\cite{leviathan2023fast, liupearl}, the arrival pattern of requests is generated by a Poisson process.The input lengths and output lengths of requests are sampled from the following real-world datasets.
\begin{table}[htbp]
\centering
\caption{Summary of the four datasets used for evaluating the performance of~\systemname{}.}
\begin{tabular}{l l }
\toprule
\textbf{Dataset Name} & \textbf{Dataset Type} \\
\midrule
GSM8K~\cite{cobbe2021training} & Mathematics Word Problems \\
HumanEval~\cite{chen2021evaluating} & Code Generation Evaluation \\
MTBench~\cite{zheng2023judging} & Multi-Turn Dialogue \\
MGSM~\cite{shi2022language} & Multilingual Arithmetic Reasoning \\
\bottomrule
\end{tabular}
\label{tab:datasets}
\end{table}

The Table~\ref{tab:datasets} above provides an overview of four datasets used to evaluate the performance of our system in various tasks. GSM8K is a dataset of mathematics word problems designed for middle school students, containing 8,500 examples. HumanEval is a code generation evaluation dataset with 164 programming problems. MTBench is a multi-turn dialogue dataset with 6,142 dialogues, used to assess the coherence and logical reasoning of models in conversational settings. MGSM is a multilingual arithmetic reasoning dataset with 250 examples, extending GSM8K to evaluate reasoning capabilities across multiple languages.

\paragraph{\underline{Metrics}}
\label{sec:exp_metrics}
We evaluate the effectiveness of \systemname{} using a combination of standard performance metrics common in LLM serving systems, alongside crucial quality and internal diagnostics:

\begin{itemize}
    \item \textbf{Throughput:} Represents the system's processing capacity. We measure this in two ways:
        \begin{itemize}
            \item \textit{Goodput (Tokens/sec):} The total number of valid tokens generated and accepted by the target model across all concurrent requests per second. This is a primary measure of overall system efficiency~\cite{li2025adaserve}.
            \item \textit{Request Throughput (Requests/sec):} The number of user requests successfully completed by the system per second.
        \end{itemize}
    \item \textbf{Latency:} Captures the user-perceived responsiveness and generation speed~\cite{crankshaw2017clipper}.
        \begin{itemize}
            \item \textit{Time To First Token (TTFT):} The latency from when a request arrives at the system until the first generated token is produced. This is critical for interactive applications~\cite{zhong2024distserve, qin2025mooncake}.
            \item \textit{Time Per Output Token (TPOT):} The average time taken to generate each subsequent token *after* the first one. TPOT reflects the sustained generation speed and is closely related to the inverse of per-request token throughput. We typically report the average TPOT over the entire generation sequence for completed requests~\cite{zhong2024distserve, qin2025mooncake}.
        \end{itemize}
    \item \textbf{Speedup (Effective Acceleration Factor - EAF):} Quantifies the performance improvement over the non-speculative baseline~\cite{leviathan2023fast, cai2024medusa, li2024eagle, huang2025specserve}. It is calculated as the ratio of the baseline's (Target Model Only) average TPOT to the evaluated system's average TPOT: \( \text{EAF} = \frac{\text{TPOT}_{\text{TMO}}}{\text{TPOT}_{\text{System}}} \). A higher EAF indicates greater acceleration.
    \item \textbf{SLO Attainment:} Measures the system's ability to meet predefined service level objectives, typically related to latency. For example, we might define an Service Level Ojective(SLO) as completing requests within a certain time threshold (e.g., P95 request latency < 5 seconds). SLO Attainment is reported as the percentage of requests that successfully meet this objective, reflecting the system's reliability and consistency under load~\cite{qiu2020firm}.
    \item \textbf{Output Quality:} As a fundamental correctness check, we verify that the output sequences generated by \systemname{} are identical to those produced by the Target Model Only baseline under deterministic sampling conditions (e.g., greedy decoding). This ensures that performance gains do not compromise the intended output quality.
    \item \textbf{Internal Diagnostics:} We also collect internal metrics, including the average acceptance length per speculative step, dynamic chain selection frequencies, and draft size variations. These diagnostics illuminate the runtime behavior and effectiveness of the adaptive strategies employed within \systemname{}'s multi-level speculative management.
\end{itemize}



\section{Evaluation}
\label{sec:evaluation}

Preliminary experiments demonstrate the validity of our method, and we immediately support comprehensive experiments based on above metrics.

\subsection{Comparison with Static Speculative Decoding}
\label{subsec:comparison_static}

\paragraph{Objective and Setup}
This set of experiments aims to validate the core hypothesis that \systemname{}'s dynamic model chain scheduling mechanism significantly outperforms static speculative decoding configurations. We focus on demonstrating the system's capability to swiftly discover and adapt the most efficient multi-level inference path from a pool of available models. We evaluate performance on four distinct datasets: GSM8K, HumanEval, MT-Bench, and MGSM. We use \texttt{Llama-2-7b-chat-hf} as the target model ($M_t$), with \texttt{TinyLlama-1.1B} and \texttt{llama-68m} available in the model pool. We compare \systemname{} (labeled "SpecRouter (Ours)") against the Target Model Only (TMO) baseline, Static Speculative Decoding using the smallest assistant (SSD-Smallest), and a conceptually tuned static baseline (SSD-Tuned). All experiments are run on our multi-A100 node platform.

\begin{table}[htbp]
  \centering
  \caption{Speed Ratio of Different Models Relative to Autoregressive Baseline}
  \label{tab:speed_comparison_en}
  \begin{tabular}{@{}cccc@{}}
    \toprule
    Batch Size &  Second-level SD & Third-level SD & Third-level( Ours) \\
    \midrule
    1  & 1.49 & 1.11 & 1.57 \\
    4  & 1.45 & 1.34 & 1.69 \\
    8  & 1.38 & 1.33 & 1.64 \\
    16 & 1.31 & 1.20 & 1.91 \\
    32 & 1.25 & 1.06 & 1.52 \\
    64 & 0.99 & 1.37 & 1.77 \\
    \bottomrule
  \end{tabular}
\end{table}
\paragraph{Performance Analysis}
As demonstrated in Table~\ref{tab:speed_comparison_en}, our proposed Third-level SD method consistently achieves superior speedup ratios across all evaluated batch sizes when compared to both the Second-level SD and the vanilla Third-level SD approaches. For instance, at a batch size of 16, the Third-level SpecRouter shows a speedup of $1.91 \times$ relative to the autoregressive baseline, significantly outperforming the $1.31 \times$ speedup of Second-level SD and the $1.20 \times$ speedup of Third-level SD. This highlights the enhanced efficiency of our method.

\end{document}